\documentclass{article}
\usepackage[final,nonatbib]{nips_2018}

\usepackage[utf8]{inputenc} 
\usepackage[T1]{fontenc}    
\usepackage{hyperref}       
\usepackage{url}            
\usepackage{booktabs}       
\usepackage{amsfonts}       
\usepackage{nicefrac}       
\usepackage{microtype}      
\usepackage{amsmath}
\usepackage{multirow}

\usepackage{graphicx}
\usepackage{amsmath}
\usepackage{amsthm}
\usepackage{amssymb}
\usepackage{mathtools}
\usepackage{color}

\newtheorem{algorithm}{Algorithm}[section]

\usepackage{times}
\usepackage{graphicx} 
\usepackage{subfigure}

\usepackage{algorithm}

\usepackage{hyperref}

\usepackage{times}
\usepackage{xcolor}
\usepackage{soul}
\usepackage[utf8]{inputenc}
\usepackage[small]{caption}

\usepackage{latexsym}
\usepackage{booktabs}
\usepackage{amsmath}
\usepackage{verbatim}
\usepackage{amssymb}
\usepackage{amsthm}
\usepackage{algorithm}
\usepackage{graphicx}
\usepackage{subfigure}
\usepackage{booktabs}
\usepackage{epstopdf}
\usepackage{lipsum}
\usepackage{color}

\usepackage[utf8]{inputenc} 
\usepackage[T1]{fontenc}    
\usepackage{hyperref}       
\usepackage{url}            
\usepackage{booktabs}       
\usepackage{amsfonts}       
\usepackage{nicefrac}       
\usepackage{microtype}      

\usepackage{multicol}

\usepackage{hyperref}
\usepackage{url}

\newcommand{\norm}[1]{\left\lVert#1\right\rVert}

\usepackage{wrapfig}

\usepackage{cleveref}

\usepackage{algorithmicx} 
\algnewcommand\STATE{\State}
\algnewcommand\algorithmicgiven{\textbf{Given:}}
\algnewcommand\GIVEN{\item[\algorithmicgiven]}
\algnewcommand\algorithmicinitialize{\textbf{Initialize:}}
\algnewcommand\INITIALIZE{\item[\algorithmicinitialize]}
\algnewcommand{\LineComment}[1]{\Statex \(\triangleright\) #1} 

\title{Continual Learning with Self-Organizing Maps}

\author{
  Pouya Bashivan, Martin Schrimpf, Robert Ajemian  \\
  McGovern Institute for Brain Research and\\
  Department of Brain and Cognitive Sciences\\ Massachusetts Institute of Technology \\
  Cambridge, MA, USA \\
  \texttt{\{bashivan,mschrimpf,ajemian\}@mit.edu} \\
  \And
  Irina Rish, Matthew Riemer, Yuhai Tu\\
IBM Research and MIT-IBM Watson AI Lab \\
  Yorktown Heights, NY, USA \\
  \texttt{\{rish,mdriemer,yuhai\}@us.ibm.com} \\
}

\begin{document}

\maketitle

\begin{abstract}
Despite remarkable successes achieved by modern neural networks in a wide range of applications, these networks perform best in domain-specific stationary environments where they are trained only once on large-scale controlled data repositories. 
When exposed to non-stationary learning environments, current neural networks tend to forget what they had previously learned, a phenomena known as \emph{catastrophic forgetting}. Most previous approaches to this problem rely on  memory replay buffers which store samples from previously learned tasks,  and  use them to regularize the learning on new ones. 
This approach suffers from the important disadvantage of not scaling well to real-life problems in which the memory requirements become enormous. We propose a memoryless method that combines standard supervised neural networks with self-organizing maps to solve the continual learning problem. The role of the self-organizing map is to adaptively cluster the inputs into appropriate task contexts -- without explicit labels -- and allocate network resources accordingly.  Thus, it selectively routes the inputs in accord with previous experience, ensuring that past learning is maintained and does not interfere with current learning. Out method is intuitive, memoryless, and performs on par with current state-of-the-art approaches on standard benchmarks.




\end{abstract}


\section{Introduction}
\vspace{-0.1in}
Learning algorithms typically operate on the entire training data set, in an offline mode, and may require costly re-training from scratch when new data becomes available. In contrast, humans learn continuously ("online"), adapting to the environment while leveraging past experiences. In this setting of {\em continual} learning, an agent is presented with a stream of samples (input-output pairs) from a non-stationary data distribution, such as a sequence of different classification tasks.
The agent has to learn the input-output relationships associated with different data distributions in order to be able to adapt well to new tasks without forgetting the tasks learned previously, i.e. to avoid the {\em catastrophic forgetting} \cite{catastrophic} problem in continual learning. Several types of approaches to this problem have been proposed in the past that introduce structural robustness \cite{goodfellow2013,srivastava2013}, regularize the parameters \cite{yosinski,Zenke2017,ewc} or utilize memory buffers \cite{gem,mbpa}. While the memory-based methods are among the most successful methods, these approaches generally utilize explicit task labels when storing instances of past learning and they also don't scale well to real life problems.

Herein, we propose to capture the essence of the multimodal data distributions encountered in continual learning via the unsupervised learning mechanism known as Self-Organizing Map (SOM).  The map learns simultaneously with a supervised feedforward neural net in such a way that the SOM routes each input sample to the most relevant part of the network.  Unlike previous methods, SOMs do not require explicit information about the change in task, or an explicit memory of previous samples, while still achieving performance levels close to the current state-of-art on several benchmarks.

\vspace{-0.1in}
\section{Self-Organizing Maps for Continual Learning}
\vspace{-0.1in}



%
\begin{wrapfigure}{R}{0.38\textwidth}
  \vspace{-0.25in}
  \begin{center}
    \includegraphics[height=1.6in,width=.38\textwidth]{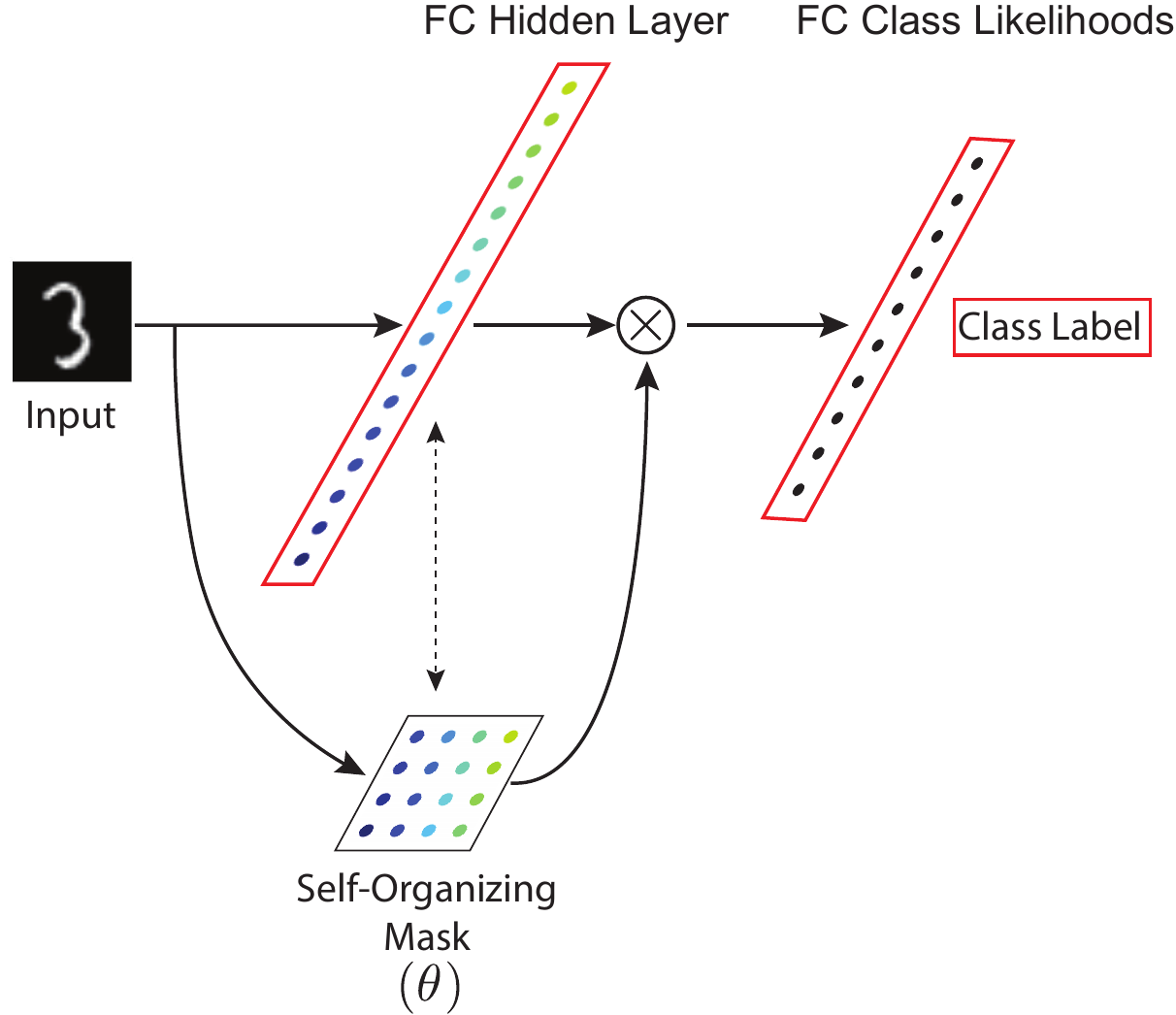}
  \end{center}
  \caption{\textbf{SOMLP setup.} 
  Red borders indicate supervised training; SOM is trained in an unsupervised manner.
  }
  \label{fig_som_setup}
  \vspace{-0.12in}
\end{wrapfigure}

Self-Organizing Maps (SOMs) \cite{Kohonen1990} are a type of artificial neural network which reduce the input from a high-dimensional ($N$) space to a low-dimensional representation without supervision.
Unlike supervised training via error-correction (backpropagation), self-organizing maps employ unsupervised competitive learning in order to map similar input vectors to physically nearby nodes in the map layer.
Namely, each SOM node will be associated with an $N$-dimensional weight vector, and, at each iteration,  the next input sample will be assigned to  the most similar network node called  the Best Matching Unit (BMU). Trainable weights associated with the BMU along with weights of its topographical neighbors (determined by neighborhood radius $\sigma$) are tuned to reduce the error between the weights and the input, (proportioanl to a learning rate $\alpha$) so that the neighborhood becomes more similar to the input. 
This allows the SOM to only adapt parts of its parameters in response to each observed input. To stabilize the learning, both $\alpha$ and $\sigma$ parameters are decayed during learning (with an exponential decaying rate of $\tau$).

SOMs has been previously used as a potential solution to the catastrophic forgetting problem \cite{Gepperth2016} but it has been shown that the proposed algorithms often fail on small-scale benchmarks like MNIST-permutations and rotations \cite{Kemker2017}. Here we introduce the Self-Organized Multi-Layer Perceptron (SOMLP) where a SOM layer is used in parallel with a fully-connected layer, as illustrated in \Cref{fig_som_setup}.
The SOM receives the same input as the fully-connected layer and is trained without supervision on the network inputs (\Cref{alg_som}). During the training it learns the input distribution for each task and, most importantly, a 2D map of relationships between the tasks. 
For every input, an output mask $\Gamma$ is computed based on the Euclidean distances between the input and SOM weight vectors, as shown in \Cref{alg_som} (tuned with hyperparameter $\epsilon$), and is multiplied with the output of the fully connected layer (\Cref{fig_som_setup}).
This allows the network to allocate resources (i.e. nodes) to learning input-output associations for each task, and mask irrelevant nodes. In addition, the SOM shares nodes between similar tasks while using separate nodes for dissimilar tasks. 
The overall training procedure is shown in \Cref{alg_training}, which includes the option of pretraining SOM weights on unlabeled data from different tasks (if available).

\begin{figure}[h]
\begin{minipage}[t]{.48\textwidth}
  \null
  \begin{algorithm}[H]
    \caption{SOM update}
    \label{alg_som}
    \begin{algorithmic}[1]
        \GIVEN input $x$ and SOM weights $\theta$ at step $t$, nodes coordinate matrix $L$, batch size $N_{bs}$, number of nodes $N_n$, 
        and hyperparameters $\alpha,\sigma,\epsilon,\tau$.

        \STATE $\Bar{x} = \frac{1}{N_{bs}}\sum_{i\in batch}^{N_{bs}}{x_i} ~~$ \Comment{average  over batch}
        
        \LineComment{Find best matching unit (BMU) for $\Bar{x}$ } 
        
        \STATE $i^*=argmin_i(\norm{\Bar{x}-\theta_i})$
        
        \STATE {\bfseries }\textbf{for } $i = 1, 2, . . . ,N_{n}$ \textbf{do} \Comment{compute}
        \STATE \quad $D_{i} = \norm{L_{i} - L_{i^*}}$  \Comment{distances to BMU}
        \STATE {\bfseries } \textbf{end} 
        \STATE $\phi = e^{D/\sigma^2}$\Comment{neighborhood mask}  
        \STATE $\Gamma=e^{-\norm{x-\theta}/\epsilon}$ \Comment{output mask}
        
        \STATE $\alpha \leftarrow \alpha e^{-\tau t /N_{steps}}$ \Comment{Adjust $\sigma$ and $\alpha$}
        \STATE $\sigma \leftarrow \sigma e^{-\tau t /N_{steps}}$
        \STATE $\theta_i \leftarrow \theta_i + \alpha \phi (x-\theta_i)$  \Comment{Update weights }
        
    \end{algorithmic}
  \end{algorithm}
\end{minipage}%
\hfill
\begin{minipage}[t]{.48\textwidth}
  \null
  \begin{algorithm}[H]
    \caption{Training procedure}
    \label{alg_training}
    \begin{algorithmic}[1]
    \GIVEN   Training datasets $(X_{t}, Y_{t})$ for each task $t$ out of $N_T$ tasks, and   data subsets $(X^{pr}_{t}, Y^{pr}_{t})$ available for SOM pretraining.
    \INITIALIZE SOM and MLP weights randomly.
     
    \LineComment{Pretrain sequentially w/ unlabeled data} 
    \STATE {\bfseries }\textbf{for } $t = 1, 2, . . . ,N_{T}$ \textbf{do}
    \STATE \quad {\bfseries }\textbf{for } batch in $ X^{pr}_{t}$ \textbf{do}
    \STATE \quad \quad Update SOM weights using Alg. \ref{alg_som}
    \STATE {\bfseries }\quad \textbf{end}
    \STATE {\bfseries } \textbf{end}
     
    \LineComment{Train sequentially using labeled data}
    \STATE {\bfseries }\textbf{for } $t = 1, 2, . . . ,N_{T}$ \textbf{do}
    \STATE \quad {\bfseries }\textbf{for } batch in $ (X_{t}, Y_{t})$ \textbf{do}
    \STATE \quad \quad Update SOM weights using Alg. \ref{alg_som}
    \STATE \quad \quad Update MLP weights using SGD
    \STATE {\bfseries }\quad \textbf{end}
    \STATE {\bfseries } \textbf{end}
    \end{algorithmic}
  \end{algorithm}
\end{minipage}
\end{figure}

\vspace{-0.1in}
\section{Experiments}
\vspace{-0.1in}
We evaluated SOMLP against three other baseline methods: a naive Multi-Layer Perceptron (MLP), Elastic Weight Consolidation (EWC) \cite{ewc} and Gradient Episodic Memory (GEM) \cite{gem}. We evaluated all methods on two standard benchmark datasets, \emph{MNIST-permutations} and \emph{MNIST-rotations} \cite{gem}. 
MNIST-permutations consists of 20 random pixel permutations of the standard MNIST dataset while MNIST-rotations contains 20 rotations of MNIST digits between 0 and 180 degrees. 

All experiments were performed on a fully-connected network with a single hidden layer with $n_{h1}$ units, varying across the experiments,  
and an output layer with $n_{h2} = 10$ units. 
For SOMLP, the hidden layer is augmented with a SOM of the same size and the output of $n_{h1}$ is multiplied by the SOM output before being fed to the next layer.
For a fair comparison with respect to the numbers of parameters, we used $n_{h1} = 3200$ for MLP, EWC and GEM and $n_{h1} = 1600$ for SOMLP\footnote{The total number of parameters for MLP, EWC and GEM is $2,540,800 = 3200 \times 784 + 3200 \times 10$; and for SOMLP $2,524,800 = 1600 \times 2 \times 784 + 1600 \times 10$}.

All networks were trained for 1 epoch on the training set, learning 20 tasks in sequence. 
For each task, the network is presented with 60,000 samples. The   hyperparameter settings used in the experiments are summarized in supplementary \Cref{table_hyperparameters}.

In addition, EWC and GEM methods require additional memory slots.
EWC uses a memory buffer to save samples before computing the Fisher matrix before moving to the next task. On the other hand,
GEM uses a memory buffer per task to retain samples for experience replay.
A scalar "memory strength" additionally dictates how much to weigh previous samples.
In our experiments, we varied the number of memory slots between 8 to 5120 for EWC with a memory strength of 3, and 256 to 5120 for GEM with a memory strength of 0.5.
An advantage of SOMLP is that it does not require a sample memory buffer.

In our approach, we used two methods for pretraining SOM weights. The first one pretrains the weights on the unlabeled MNIST training set, with large neighborhood values ($\textnormal{SOMLP}_s$), while for the second one, the weights are pretrained on 10\% of the training set from all the tasks in the benchmark ($\textnormal{SOMLP}_m$). We only use the input images for pretraining. 

\vspace{-0.1in}
\section{Results}
\vspace{-0.1in}

\Cref{table_perfs} shows the networks' performances on each benchmark dataset after learning a battery of 20 tasks.
\begin{table}[bh]
\begin{center}
\begin{small}
    \caption{Comparison of average performances on benchmark datasets. Number of parameters in each network were matched by changing the network size.}
    \resizebox{\textwidth}{!}{
    \begin{tabular}{c|c|c|c|c}
    \toprule
    \label{table_perfs}
    \textbf{Network} & \textbf{Memory Size} & \textbf{\#Parameters (M)} & \textbf{\begin{tabular}[c]{@{}c@{}}Performance (\%)\\ (MNIST-Permutations)\end{tabular}} & \textbf{\begin{tabular}[c]{@{}c@{}}Performance (\%)\\ (MNIST-Rotations)\end{tabular}} \\
    \midrule
    MLP              & 0                 & 2.5           & 29$\pm$0.6                           & 64.4$\pm$0.4         \\
    EWC-Low Mem      & 8                  & 2.5           & 84.8                              & 59.1             \\
    EWC-High Mem     & 5120             & 2.5           & 83.6                                 & 49.8             \\
    GEM-Low Mem      & 320                 & 2.5           & 68.3                             & 71.5             \\
    GEM-High Mem     & 5120            & 2.5                  & 93.3                           & 92.6             \\
    $\textnormal{SOMLP}_s$         & 0         & 2.5          & 65.9$\pm$5.2                      & 59.1$\pm$6.6         \\
    $\textnormal{SOMLP}_m$      & 0         & 2.5          & 85.4$\pm$1.8                      & 79.1$\pm$0.8      \\
    \bottomrule
    \end{tabular}}
    \end{small}
\end{center}
\end{table}
On MNIST-permutations, $\textnormal{SOMLP}_m$ performed on par with EWC but lower than GEM with large memory buffers, whereas it outperformed EWC on MNIST-rotations but still was lower than GEM. On the other hand, in the low-memory case that the size of available memory is reduced, $\textnormal{SOMLP}_m$ performs better than all baseline methods. In contrast, $\textnormal{SOMLP}_s$ only performed better than EWC on MNIST-rotations. Naive MLP worked fairly well on MNIST-rotations but lower than all methods on MNIST-permutations.


\begin{figure}[t]
\center
\includegraphics[width=\linewidth]{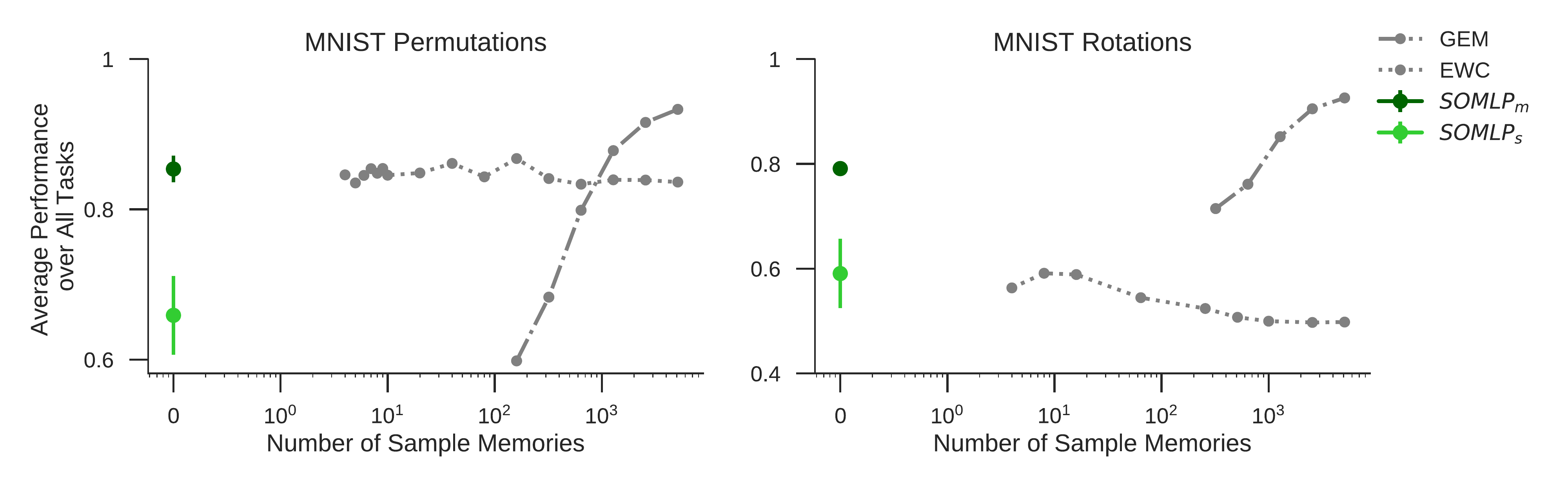}
\caption{\textbf{Sample memory size effect on performance.}
Both GEM and EWC depend on the right amount of memory.
For GEM, more memory generally leads to better performance, whereas the optimal EWC memory size seems to depend on the dataset.
SOMLP does not require sample memory.
}
\label{fig_mem}
\end{figure}
\begin{figure}[h]
\center
\includegraphics[width=.7\linewidth]{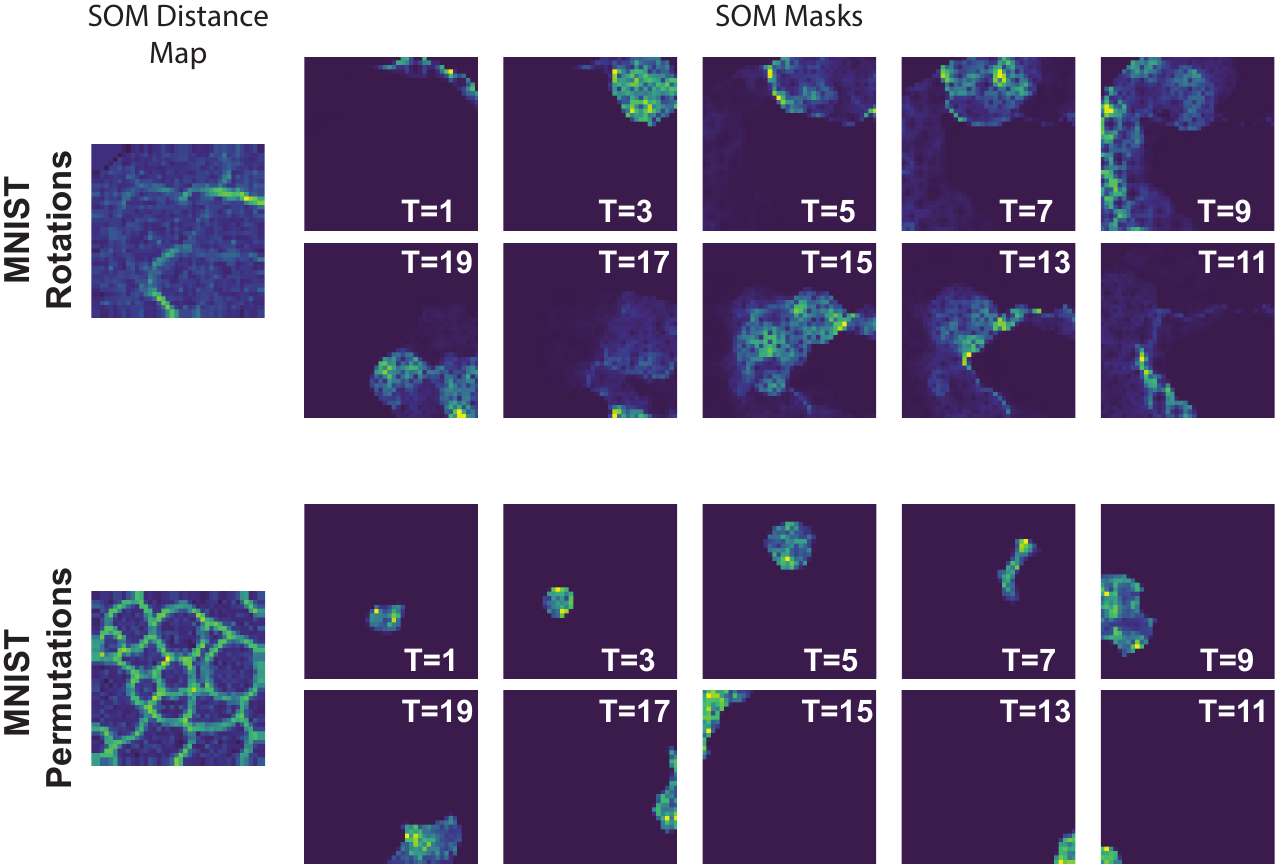}
\caption{\textbf{Learned SOM.}
(left) Euclidean distance between SOM nodes for each benchmark during a single run. (right) SOM masks found for different tasks during a single run on each benchmark.}
\label{fig_masks}
\end{figure}

It should be noted that both EWC and GEM make use of a "sample memory buffer" as well as explicit information about when the task is switched while SOMLP requires neither a memory buffer nor knowledge of task switching.
We examined the effect of the amount of sample memory on mean performance\footnote{We note that EWC additionally stores the Fisher matrix associated with each task which we did not account for in our memory requirement estimate.}. 
We found that the performance of GEM on both benchmarks is significantly reduced with smaller memory sizes. EWC performance is almost constant in the MNIST-permutations benchmark and even decreases with enlarged memory buffer in MNIST-rotations (\Cref{fig_mem}). 

We also examined the mean performance during sequential learning of all tasks (\Cref{fig_pertask}) on each dataset. As more tasks are learned, EWC and $\textnormal{SOMLP}_s$ experience some degree of forgetting and therefore the mean performance decreases with more tasks. While both GEM and $\textnormal{SOMLP}_m$ are able to maintain the same level of mean performance throughout, $\textnormal{SOMLP}_m$ is consistently below GEM's level of performance. This is potentially due to lower network capacity in our SOMLP approach because of (1) smaller network size and (2) per-task SOM masks that limits the resources available to learn each task. 

The learned feature maps and per-task masks for each benchmark are shown in \Cref{fig_masks}. 
In MNIST-permutations, because of the random pixel permutations in each task, masks corresponding to each task are independent of each other. Conversely, in MNIST-rotations, the learned masks share nodes between tasks that are more similar (i.e. slightly rotated tasks) and use independent nodes for more dissimilar tasks (e.g. tasks that are separated by larger rotation angles).

\textbf{Acknowledgements}:
This research was supported by the MIT-IBM Watson AI Lab and the Semiconductor Research Corporation (SRC). 

\bibliography{SOM_continual_full}
\bibliographystyle{plain}

\renewcommand\thetable{S\arabic{table}}  
\renewcommand\thefigure{S\arabic{figure}}  
\clearpage
\section*{Supplementary Material}
\setcounter{table}{0}
\setcounter{figure}{0}

\begin{table}[bh]
\begin{center}
    \caption{List of hyperparameter choices for SOMLP and baseline method.}
    \label{table_hyperparameters}
    \resizebox{\textwidth}{!}{
    \begin{tabular}{c|c|c|c|c|c|c|c|c}
    \toprule
    \multirow{2}{*}{\textbf{Hyperparameter}} & \multicolumn{2}{c|}{$\textnormal{SOMLP}_s$}            & \multicolumn{2}{c|}{$\textnormal{SOMLP}_m$}        &      \multicolumn{2}{c|}{\textbf{EWC}}            &
    \multicolumn{2}{c}{\textbf{GEM}}    \\ 
    & Permutations         & Rotations            & Permutations         & Rotations       & Permutations         & Rotations            & Permutations         & Rotations     \\ \midrule
    $\alpha$                          & 0.5                  & 3                    & 1                    & 3               & -   & -      & -      & -    \\ 
    $\sigma$                          & 4                    & 4                    & 2                    & 3                & -   & -       & -      & -    \\
    $\epsilon$                        & 0.5                  & 0.5                  & 0.5                  & 0.5               & -   & -      & -      & -      \\
    $\tau$                            & 2                    & 3                    & 15                   & 3                 & -   & -      & -      & -   \\
    Batch Size                      & 10                    & 100                  & 10                   & 100              & 10   & 10     & 10      & 10   \\
    Learning Rate                   & 0.01                    & 0.02                  & 0.01                   & 0.02              & 0.1   & 0.01      & 0.1       & 0.1 \\
    \bottomrule
    \end{tabular}
    }
\end{center}
\end{table}


\begin{figure}[h]
\center
\includegraphics[width=\linewidth]{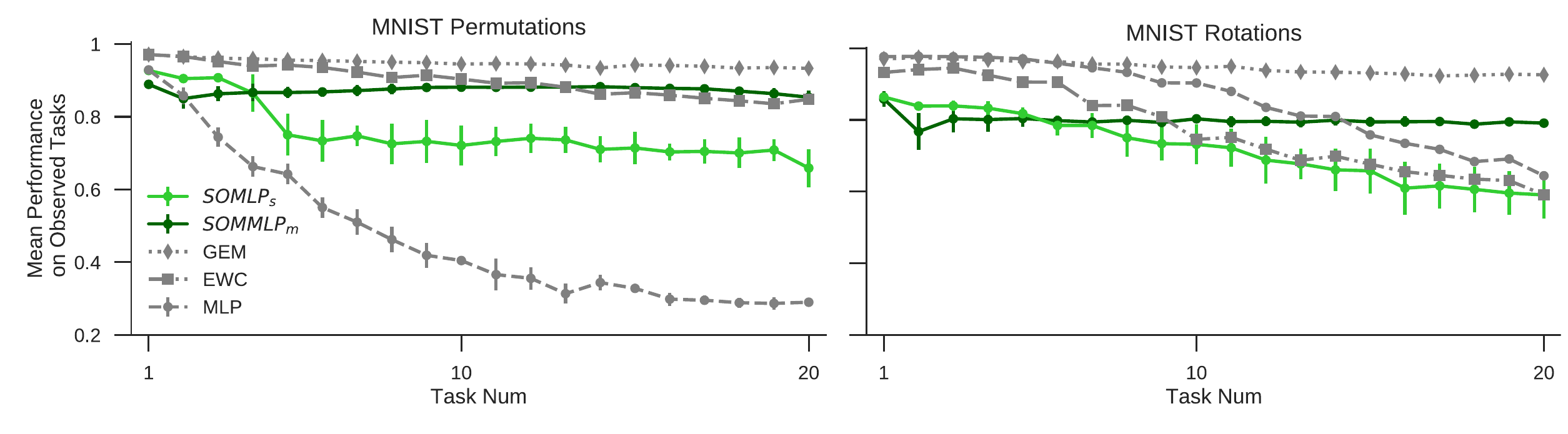}
\caption{\textbf{Performance during task learning.}
Average performance on observed tasks during sequential learning of different task-sets decreases for a standard MLP, while continual learning methods maintain performance with varying degrees of success.}
\label{fig_pertask}
\end{figure}

\end{document}